\documentclass[12pt]{elsarticle}
\usepackage{times}
\usepackage{epsfig}
\usepackage{graphicx}
\usepackage{amsmath}
\usepackage{amssymb}
\usepackage{amsthm}
\usepackage{makeidx}  
\usepackage{url}
\usepackage{algorithm}
\usepackage{bm}
\usepackage{microtype}
\usepackage{graphicx}
\usepackage{url}
\usepackage{caption}
\usepackage{subfig}
\usepackage{tabularx}
\usepackage[rightcaption]{sidecap}
\usepackage{pdflscape}
\usepackage[bookmarks=true]{hyperref}
\usepackage{lineno}
\usepackage{ulem}
\usepackage{afterpage}
\usepackage[usenames,dvipsnames]{color}

\newcommand{\bPsi}{\bm{\Psi}}

\newcommand{\bOmega}{\bm{\Omega}}

\newcommand{\DSI}{S}

\newcommand{\Do}{\mathbb{S}}

\renewcommand{\Re}{{\mathbb{R}}}
\newcommand{\ie}{{i.e., }}

\newcommand{\Id}{{\mathtt{Id}}}

\newcommand{\y}{{\bf y}}

\newcommand{\bL}{\mathbf{L}}

\newcommand{\tr}{\text{trace}}

\renewcommand{\t}{{\bf t}}

\newcommand{\M}{{\bf M}}

\newcommand{\rb}{{\bf r}}
\newcommand{\x}{{\bf x}}
\newcommand{\s}{{\bf q}}

\newcommand{\bmu}{\bm{\mu}}
\newcommand{\ww}{\bm{\omega}}

\newcommand{\cc}{{\bf c}}
\newcommand{\dd}{{\bf c^{(s)}}}

\newcommand{\ub}{\nabla_\x h}

\newcommand{\atlas}{\text{atlas}}

\newcommand{\old}{\text{old}}
\newcommand{\new}{\text{new}}
\DeclareMathOperator*{\argmax}{arg\,max}
\DeclareMathOperator*{\argmin}{arg\,min}

\floatstyle{ruled}
\newfloat{alg}{thp}{lop}
\floatname{alg}{Algorithm}
\renewenvironment{algorithm}
{\begin{alg} \vspace{0cm} \noindent } {\vspace{0cm}
\end{alg}}

\journal{NeuroImage}

\begin{document}

\title{Diffeomorphic Metric Mapping and Probabilistic Atlas Generation of Hybrid Diffusion Imaging based on BFOR Signal Basis}


%

\author[l1]{Jia Du}
\author[|2,|3]{A. Pasha Hosseinbor}
\author[|3,|4]{Moo K. Chung}
\author[|5]{Barbara B. Bendlin}
\author[|3]{Gaurav Suryawanshi}
\author[|2,|3]{Andrew L. Alexander}
\author[l1,l6,l7]{Anqi Qiu\corref{cor1}}
\ead{bieqa@nus.edu.sg}

\cortext[cor1]{Correspondence to: Anqi Qiu, Department of Bioengineering,
National University of Singapore, 9 Engineering Drive 1, Block EA
03-12, Singapore 117576. Tel: +65 6516
7002. Fax: +65 6872 3069}
\address[l1]{Department of Bioengineering, National University of Singapore, Singapore}
\address[|2]{Department of Medical Physics, University of Wisconsin-Madison, USA}
\address[|3]{Waisman Laboratory for Brain Imaging and Behavior, University of Wisconsin-Madison, USA}
\address[|4]{Biostatistics and Medical Informatics, University of Wisconsin-Madison, USA}
\address[|5]{Department of Medicine, University of Wisconsin-Madison, USA}
\address[l6]{Singapore Institute for Clinical Sciences, Agency for Science, Technology and Research, Singapore}
\address[l7]{Clinical Imaging Research Center, National University of Singapore, Singapore}

\begin{abstract}
We propose a large deformation diffeomorphic metric mapping algorithm to align multiple $b$-value diffusion weighted imaging (mDWI) data, specifically acquired via hybrid diffusion imaging (HYDI), denoted as LDDMM-HYDI. We then propose a Bayesian model for estimating the white matter atlas from HYDIs. We adopt the work given in Hosseinbor et al. (2012) and represent the $q$-space diffusion signal with the Bessel Fourier orientation reconstruction (BFOR) signal basis. The BFOR framework provides the representation of mDWI in the $q$-space and thus reduces memory requirement. In addition, since the BFOR signal basis is orthonormal, the $\bL^2$ norm that quantifies the differences in the $q$-space signals of any two mDWI datasets can be easily computed as the sum of the squared differences in the BFOR expansion coefficients. In this work, we show that the reorientation of the $q$-space signal due to spatial transformation can be easily defined on the BFOR signal basis. We incorporate the BFOR signal basis into the LDDMM framework and derive the gradient descent algorithm for LDDMM-HYDI with explicit orientation optimization. Additionally, we extend the previous Bayesian atlas estimation framework for scalar-valued images \cite{jun_nimg_2008} to HYDIs and derive the expectation-maximization algorithm for solving the HYDI atlas estimation problem. Using real HYDI datasets, we show the Bayesian model generates the white matter atlas with anatomical details. Moreover, we show that it is important to consider the variation of mDWI reorientation due to a small change in diffeomorphic transformation in the LDDMM-HYDI optimization and to incorporate the full information of HYDI for aligning mDWI.
\end{abstract}

\maketitle

\section{Introduction}

Diffusion-weighted MRI methods are promising tools for characterizing tissue microstructure.  While diffusion tensor imaging (DTI) and high angular resolution diffusion imaging (HARDI) methods are widely used methods, they do not provide a complete description of the diffusion distribution. In order to more accurately reconstruct the ensemble average propagator (EAP), a thorough sampling of the $q$-space sampling is needed, which requires multiple $b$-value diffusion weighted imaging (mDWI).  The EAP estimation using mDWI better characterizes more complex neural fiber geometries and non-Gaussian diffusion behavior when compared to single $b$-value techniques. Recently, new $q$-space imaging techniques, diffusion spectrum imaging (DSI) \cite{Wedeen_DSI} and hybrid diffusion imaging (HYDI) \cite{Wu_Alexander_HYDI} have been developed for estimating the EAP. HYDI is a mDWI technique that samples the diffusion signal along concentric spherical shells in the $q$-space, with the number of encoding directions increased with each shell to increase the angular resolution with the level of diffusion weighting. Originally, HYDI employed the fast Fourier transform (FFT) to reconstruct the EAP. However, the recent advent of analytical EAP reconstruction schemes, which obtain closed-form expressions of the EAP, obviate the use of the FFT in HYDI. One such technique successfully validated on HYDI datasets is Bessel Fourier orientation reconstruction (BFOR) \cite{Pasha_NI2012}. While mDWI techniques like HYDI have not been widely used, the new human connectome project \cite{VanEssen201362} and spin-off projects will likely significantly increase the application. However, there is a lack of fundamental image analysis tools for mDWI and EAP maps, such as registration and atlas generation, that can fully utilize their information.


In the last decades, researchers have spent great efforts on developing registration algorithms to align diffusion tensors derived from DTI and orientation distribution functions (ODFs) derived from HARDI [e.g., \cite{Alexander:TMI01,Raffelt:Neuro2011,Du_TMI2012}]. However, registration algorithms directly based on DWIs are few. The direct alignment of DWIs in the  $q$-space utilizes the full diffusion information, is independent of the choice of diffusion models and their reconstruction algorithms (e.g., tensor, ODF), and unifies the transformation to align the local diffusion profiles defined at each voxel of two brains \cite{Dhollander_MICCAI2010,Yap_TMI2012,zhang_miccai2012}. Dhollander et al.\cite{Dhollander_MICCAI2010} developed an algorithm that transforms the diffusion signals on a single shell of the $q$-space and preserves anisotropic as well as isotropic volume fractions. Yap et.al \cite{Yap_TMI2012} proposed to decompose the diffusion signals on a single shell of the $q$-space into a series of weighted diffusion basis functions, reorient these functions independently based on a local affine transformation, and then recompose the reoriented functions to obtain the final transformed diffusion signals. This approach provides the representation of the diffusion signal and also explicitly models the isotropic component of the diffusion signals to avoid undesirable artifacts during the local affine transformation. Zhang et al. \cite{zhang_miccai2012} developed a diffeomorphic registration algorithm for aligning DW signals on a single shell of the $q$-space. 

Only recently, Dhollander et al. \cite{Dhollander_MICCAI2011} aligned DWIs on multiple shells of the $q$-space by first estimating transformation using a multi-channel diffeomorphic mapping algorithm, in which generalized fractional anisotrophy (GFA) images computed from each shell were used as mapping objects, and then applying the transformation to DWIs in each shell using the DWI reorientation method in \cite{Dhollander_MICCAI2010}. This approach neglected possible influences of the DWI reorientation on the optimization of the spatial transformation. Hsu et al. \cite{Hsu_NI2012} generalized the large deformation diffeomorphic metric image mapping algorithm \cite{Beg:LDDMM} to DWIs in multiple shells of the $q$-space and considered the image domain and $q$-space as the spatial domain where the diffeomorphic transformation is applied to. The authors claimed that the reorientation of DWIs is no longer needed as the transformation also incorporates the deformation due to the shape differences in the  diffusion profiles in the $q$-space. It is a robust registration approach with the explicit consideration of the large deformation in both the image domain and the $q$-space. However, its computational complexity and memory requirement are high. 

While limited research has been done for aligning the HYDI images, efforts on the white matter atlas from HYDI is even less. Only recently, Dhollander et al. \cite{Dhollander_MICCAI2011} employed their multi-channel diffeomorphic matching algorithm for the atlas generation using HYDI datasets. To our best knowledge, there is no research on probablistic atlas genertion for HYDI.

In this paper, we propose a new large deformation diffeomorphic metric mapping (LDDMM) algorithm to align HYDI datasets, denoted as LDDMM-HYDI, and then develop a Bayesian estimation framework for generating the brain atlas. In particular, we adopt the BFOR framework in representing the $q$-space signal \cite{Pasha_NI2012}. Unlike the diffeomorphic mapping of mDWIs in Hsu et al. \cite{Hsu_NI2012}, the BFOR signal basis provides the representation of the $q$-space signal and thus reduces memory requirement. In addition, since the BFOR signal basis is orthonormal, the $\bL^2$ norm that quantifies the differences in the $q$-space signals can be easily computed as the sum of the squared differences in the BFOR expansion coefficients. In this work, we will show that the reorientation of the $q$-space signal due to spatial transformation can be easily defined on the BFOR signal basis. Unlike the work in \cite{Dhollander_MICCAI2011}, we will incorporate the BFOR signal basis into the LDDMM framework and derive the gradient descent algorithm for solving the LDDMM-HYDI variational problem with explicit orientation optimization. Using this registration approach, we will further estimate the brain white matter atlas from the $q$-space based on a Bayesian model. This probabilistic model is the extension of the previous Bayesian atlas  estimation for scalar-based intensity images \cite{jun_nimg_2008}. With the aids of the BFOR representation and reorientation of mDWIs introduced in this work, we show that it is feasible to adopt the previous Bayesian atlas estimation model for scalar-valued images \cite{jun_nimg_2008} to HYDI.  As shown below, the main contributions of this paper are:
\begin{enumerate}
\item  to seek large deformation for aligning HYDI datasets based on the BFOR representation of mDWI.
  \item to derive  the rotation-based reorientation of the $q$-space signal via the BFOR signal basis. This is equivalent to applying Wigner matrix to the BFOR expansion coefficients, where Wigner matrix can be easily constructed by the rotation matrix (see Section \ref{sec:LocalAffine}).
\item to derive the gradient descent algorithm for the LDDMM-HYDI variational problem with the explicit orientation optimization. In particular, we provide a computationally efficient method for calculating the variation of Wigner matrix due to the small variation of the diffeomorphic transformation (see Section \ref{sec:gradODF}).
  \item to show that the LDDMM-HYDI gradient descent algorithm does not involve the calculation of the BFOR signal bases and hence avoids the discretization in the $q$-space. 
\item to propose a Bayesian estimation model for the $q$-space signals represented via the BFOR signal basis and derive an expectation-maximization algorithm for solving it (see Section \ref{sec:atlas}).
\end{enumerate}

\section{Review: BFOR Signal Basis}
\label{sec:review}
According to the work in \cite{Pasha_NI2012}, the $q$-space diffusion signal, $\DSI(\x, \s)$, can be represented as 
\begin{align}
\label{eq:HYDI_rep}
\DSI(\x,\s) &=\sum^{N_b}_{n=1}\sum^{N_Y}_{j=1}c_{nj} (\x)\Psi_{nj}(\s) \ ,
\end{align}
where $\x$ and $\s$ respectively denote the image domain and $q$-space. $\Psi_{nj}(\s)$ is the $nj$-th BFOR signal basis with its corresponding coefficient, $c_{nj}(\x)$, at $\x$. $\Psi_{nj}(\s)$ is given as 
\begin{align}
\label{eq:BSH}
\Psi_{nj}(\s)= \frac{2\sqrt{\alpha_{nl(j)}}}{\sqrt{\pi \tau^3} J_{l(j)+3/2}(\alpha_{nl(j)})}  j_{l(j)}\Big(\frac{\alpha_{nl(j)}|\s|}{\tau}\Big)Y_{j}\Big(\frac{\s}{|\s|}\Big) \ .
\end{align}
Here, $\alpha_{nl}$ is the $n^{th}$ root of the $l^{th}$ order spherical Bessel (SB) function of the first kind $j_{l}$. $\tau$ is the radial distance in $q$-space at which the Bessel function goes to zero. $Y_j$ are the modified real and symmetric spherical harmonics (SH) bases as given in \cite{Goh:MICCAI2009}. $J_{l(j)+3/2}(\cdot)$ is the Bessel function of the first kind. $N_Y=\frac{(L+1)(L+2)}{2}$ is the number of terms in the modified SH bases of truncation order $L$, while $N_b$ is the truncation order of radial basis. Note that in the original publication on BFOR \cite{Pasha_NI2012}, the BFOR basis was not normalized to unity, but we have rectified this in Eq. (\ref{eq:BSH}). We refer readers to \cite{Pasha_NI2012} for more details on the BFOR algorithm.

Using the fact that the BFOR signal basis is orthonormal, the $\bL^2$-norm of $\DSI(\x, \s)$ can be easily written as 
\begin{align}
\label{eq:DSI_norm}
\left\|\DSI(\x,\s)\right\|_{2}=\sqrt{\int_{\x\in \Re^3}\int_{\s\in\Re^3}\DSI^2(\x,\s)d\s d\x}=\sqrt{\int_{\x\in \Re^3}\sum^{N_b}_{n=1}\sum^{N_Y}_{j=1}c_{nj}(\x)^2 d\x}  \ .
\end{align}


\section{Large Deformation Diffeomorphic Metric Mapping for HYDI}
\label{sec:LDDMM-HYDI}
\subsection{Rotation-Based Reorientation of $\DSI(\x, \s)$}
\label{sec:LocalAffine}
We now discuss the reorientation of $\DSI(\x,\s)$ when rotation transformation $R$ is applied. We assume that the diffusion profile in each shell of the $q$-space remains in the same shell after the reorientation. However, its angular profile in each shell of the $q$-space is transformed according to the rotation transformation.  Hence, we define 
$$
 R \DSI(\x,\s) = \DSI\Big(\x,|\s| R^{-1}\frac{\s}{|\s|}\Big) \ .
$$
According to the BFOR representation of $\DSI(\x,\s)$ in Eq. (\ref{eq:HYDI_rep}), we thus have 
$$
 R \DSI(\x,\s) =\sum^{N_b}_{n=1}\sum^{N_Y}_{j=1}c_{nj} (\x) \frac{2\sqrt{\alpha_{nl(j)}}}{\sqrt{\pi \tau^3} J_{l(j)+3/2}(\alpha_{nl(j)})}  j_{l(j)}\Big(\frac{\alpha_{nl(j)}|\s|}{\tau}\Big)Y_{j}\Big(R^{-1}\frac{\s}{|\s|}\Big) \ .
$$
This indicates that the rotation reorientation of mDWI is equivalent to applying the rotation transformation to the real spherical harmonics, $Y_{j}$. According to the work in \cite{Geng:TMI11}, the rotation of $Y_{j}$ can be achieved by the rotation of their corresponding coefficients, yielding 
\begin{align}
\label{eqn:reorient}
R \DSI(\x,\s) =\sum^{N_b}_{n=1}\Big(\sum^{N_Y}_{j=1} \big(\sum_{j^\prime = 1}^{N_Y}M_{jj^\prime}c_{nj^\prime} (\x) \big)\Big) \frac{2\sqrt{\alpha_{nl(j)}}}{\sqrt{\pi \tau^3} J_{l(j)+3/2}(\alpha_{nl(j)})}  j_{l(j)}\Big(\frac{\alpha_{nl(j)}|\s|}{\tau}\Big)Y_{j}\Big(\frac{\s}{|\s|}\Big) \ ,
\end{align}
where $M_{jj^\prime}$ is the $jj^\prime$th element of Wigner matrix $M(R)$ constructed based on $R$ (see details in \cite{Geng:TMI11}). We can see that the same Wigner matrix is applied to $c_{nj}$ at a fixed $n$.  For the sake of simplicity, we rewrite Eq. (\ref{eqn:reorient}) in the matrix form, \ie
$$
 R \DSI(\x,\s) = \big(\M(R) \ \cc(\x)\ \big)^\top \bPsi(\s) \ , 
$$
where $\M$ is a sparse matrix with $N_b$ diagonal blocks of $M(R)$. $\cc$ is a vector that concatenates coefficients $c_{nj^\prime}$ in the order such that at a fixed $n$, $c_{nj^\prime}$ corresponds to $M(R)$. $\bPsi(\s)$ concatenates the BFOR signal basis.

\subsection{Diffeomorphic Group Action on $\DSI(\x, \s)$}
\label{sec:DiffGrpAct}
We define an action of diffeomorphisms $\phi : \Omega \rightarrow \Omega$ on $\DSI(\x,\s)$, which takes into consideration of the reorientation in the $q$-space as well as the transformation of the spatial volume in $\Omega$. Based on the rotation reorientation of $\DSI(\x, \s)$ in Eq. \eqref{eqn:reorient}, for a given spatial location $\x$, the action of $\phi$ on $\DSI(\x,\s)$ can be defined as 
\begin{align*}
\phi\cdot\DSI(\x,\s) &= \DSI\big(\phi^{-1}(\x),R^{-1}_{\phi^{-1}(\x)}\s\big) \nonumber\\ \nonumber
&=\Big(\M\big(R_{\phi^{-1}(\x)}\big)\;\cc\big(\phi^{-1}(\x)\big)\Big)^{\top} \bPsi(\s) \ ,
\end{align*}
where $R_\x$ can be defined in a way similar to the finite strain scheme used in DTI registration \cite{Alexander:TMI01}. That is, $R_\x=(D_\x\phi D_\x^{\top}\phi)^{-\frac{1}{2}}D_\x\phi$, where $D_{\x}\phi$ is the Jacobian matrix of $\phi$ at $\x$. For the remainder of this paper, we denote this as 
\begin{align}
\label{eqn:diffeoaction}
\phi\cdot\DSI(\x,\s) &= \left(\big(\M(R_{\x})\;\cc(\x)\big)^{\top} \right)\circ\phi^{-1}(\x) \;\;  \bPsi(\s)\ ,
\end{align}
where $\circ$ indicates as the composition of diffeomorphisms.

\subsection{Large Deformation Diffeomorphic Metric Mapping for HYDIs}
\label{sec:LDDMMDSI}
The previous sections equip us with an appropriate representation of HYDI mDWI and its diffeomorphic action. Now, we state a variational problem for mapping HYDIs from one subject to another. We define this problem in the ``large deformation" setting of Grenander's group action approach for modeling shapes, that is, HYDI volumes are modeled by assuming that they can be generated from one to another via flows of diffeomorphisms $\phi_t$, which are solutions of ordinary differential equations $\dot \phi_t = v_t (\phi_t), t \in [0,1],$ starting from the identity map $\phi_0={\Id}$. They are therefore characterized by time-dependent velocity vector fields $v_t, t \in [0,1]$. We define a metric distance between a HYDI volume of a subject $\DSI^{(s)}$ and an atlas  HYDI volume $\DSI^{\atlas}$ as the minimal length of curves $\phi_t \cdot \DSI^{\atlas}, t \in [0,1],$ in a shape space such that, at time $t=1$,  $\phi_1 \cdot \DSI_{\t} = \DSI^{(s)}$. Lengths of such curves are computed as the integrated norm $\| v_t \|_V$ of the vector field generating the transformation, where $v_t \in V$, where $V$ is a reproducing kernel Hilbert space with kernel $k_V$ and norm $\| \cdot \|_V$. To ensure solutions are diffeomorphic, $V$ must be a space of smooth vector fields. Using the duality isometry in Hilbert spaces, one can equivalently express the lengths in terms of $m_t$, interpreted as momentum such that for each $u\in V$, $\langle m_t, u \circ \phi_t\rangle_2 = \langle k_V^{-1}v_t, u\rangle_2$, where we let $\langle m, u\rangle_2$ denote the $\bL^2$ inner product between $m$ and $u$, but also, with a slight abuse, the result of the natural pairing between $m$ and $v$ in cases where $m$ is singular (e.g., a measure). This identity is classically written as $\phi_t^* m_t =k_V^{-1} v_t$, where $\phi_t^*$ is referred to as the pullback operation on a vector measure, $m_t$. Using the identity $\|v_t\|_V^2 = \langle k_V^{-1}v_t, v_t\rangle_2=\langle m_t,k_Vm_t\rangle_2$ and the standard fact that energy-minimizing curves coincide with constant-speed length-minimizing curves, one can obtain the metric distance between the template and target volumes by minimizing $\int_0^1 \langle m_t, k_Vm_t\rangle_2 dt$ such that $\phi_1 \cdot \DSI^{\atlas}=\DSI^{(s)} $ at time $t=1$. We associate this with the variational problem in the form of  
\begin{eqnarray}
\label{eqn:metric} 
J(m_t) =& \inf_{m_t: \dot \phi_t =
k_Vm_t(\phi_t),\phi_0=\Id} \int_0^1 \langle m_t, k_Vm_t\rangle_2 dt + \lambda \; E(\phi_1 \cdot \DSI^{\atlas},\DSI^{(s)}),
\end{eqnarray}
where $\lambda$ is a positive scalar. $E$ quantifies the difference between the deformed atlas $\phi_1 \cdot \DSI^{\atlas}$ and the subject  $\DSI^{(s)}$. Based on Eq. \eqref{eq:DSI_norm} and \eqref{eqn:diffeoaction}, $E$ is expressed in the form of 
\begin{eqnarray}
\label{eqn:Ex} 
E = \int_{\x\in\Omega}\left\| \big(\M(R_{\x})\ \cc^{\atlas}(\x) \big)\circ\phi^{-1}(\x)-\cc^{(s)}(\x)\right\|_2^2 d\x  \ . 
\end{eqnarray}

\subsection{Gradient of $J$ with respect to $m_t$}
\label{sec:gradODF}
We now solve the optimization problem in Eq. \eqref{eqn:metric} via a gradient descent method. The gradient of $J$ with respect to $m_t$ can be computed via studying a variation ${m}_t^\epsilon=m_t + \epsilon \widetilde{m}_t $ on $J$ such that the derivative of $J$ with respect to $\epsilon$ is expressed in function of $ \widetilde{m}_t$. According to the general LDDMM framework derived in \cite{Du:SD}, we directly give the expression of 
the gradient of $J$ with respect to $m_t$ as 
\begin{eqnarray}
\label{eqn:gradJ}
\nabla J(m_t) &= & 
 2 m_t + \lambda \eta_t \ ,
\end{eqnarray}
where 
\begin{equation}
\label{eqn:eta}
\eta_t = \nabla_{\phi_1}E +  \int_t^1 \bigl[ \partial_{\phi_s} (k_V m_s) \bigr]^\top (\eta_s + m_s) ds \ ,
\end{equation}
Eq. \eqref{eqn:eta} can be solved backward given $\eta_1= \nabla_{\phi_1}E$. $\;\partial_{\phi_s} (k_V m_s)$ is the partial derivative of $k_Vm_s$ with respect to $\phi_s$. 
 
In the following, we discuss the computation of $\nabla_{\phi_1} E$. We consider a variation of $\phi_1$ as  $\phi_1^\epsilon=\phi_1 + \epsilon h $ and denote the corresponding variation in $\M(R_\x)$ as  $\M(R_\x^\epsilon)$.  Denote $\hat{\cc}(\x)=\M(R_\x)\cc^{\atlas}(\x)$ for the simplicity of notation. We have   
\begin{align}
\label{eq:Egrad}
\frac{\partial E}{\partial\epsilon}\Big|_{\epsilon=0} &=\int_{\x\in\Omega}\frac{\partial \left\|(\M(R^{\epsilon}_\x)\cc^{\atlas}(\x))\circ(\phi^{\epsilon}_1)^{-1}(\x)-\cc^{(s)}(\x)\right\|_2^2}{\partial\epsilon}\Big|_{\epsilon=0}d\x \\
&=\underbrace{2\int_{\x\in\Omega}\Big\langle\hat{\cc}(\x)\circ \phi_1^{-1}-\cc^{(s)}(\x), \nabla_\x^\top \hat{\cc}(\x) \circ \phi_1^{-1} \frac{\partial (\phi^{\epsilon}_1)^{-1} }{\partial\epsilon}\Big|_{\epsilon=0} \Big\rangle d\x}_{\text{term (A)}} \nonumber \\
&+\underbrace{2\int_{\x\in\Omega} \Big\langle\hat{\cc}(\x)\circ \phi_1^{-1}-\cc^{(s)}(\x), \left(\frac{\partial \M(R_\x^{\epsilon})\cc^{\atlas}(\x)}{\partial\epsilon}\Big|_{\epsilon=0} \right)\circ \phi_1^{-1} \Big\rangle d\x}_{\text{term (B)}} \nonumber \ .
\end{align}
 
As the calculation of Term (A) is straightforward, we directly give its expression, \ie
\begin{equation}
\label{eq:termA}
\text{Term (A)} =-2\int_{\x\in\Omega} \Big\langle \big(D_\x\phi_1\big)^{-\top}\nabla_{\x} \hat{\cc}(\x)\Big(\hat{\cc}(\x)-\cc^{(s)}\big(\phi_1(\x)\big)\Big) \det{\big(D_\x\phi_1\big)}, h \Big\rangle d\x \ .
\end{equation}
This term is similar to that in the scalar image mapping case. It seeks the optimal spatial transformation $\phi_t$ in the gradient direction of image $\hat{\cc}(\x)$ weighted by the difference between the atlas and subject's images.

The computation of Term (B) involves the differential of $\M(R_\x)$ with respect to rotation matrix $R_\x$ and the variation of $R_\x^\epsilon$ with respect to the small variation of $\phi_1^\epsilon$. Let's first compute the derivative of $\M(R_\x)$ with respect to rotation matrix $R_\x$. According to the work in \cite{Cetingul_ISBI2012}, the analytical form of this derivative can be solved using the Euler angle representation of $R_\x$ but is relatively complex. Here, we consider Wigner matrix $\M(R_\x)$ and the coefficients of the BFOR signal basis $\cc^{\atlas}(\x)$ together, which leads to a simple numeric approach for computing the derivative of $\hat{\cc}(\x)=\M(R_\x)\cc^{\atlas}(\x)$ with respect to rotation matrix $R_\x$, \ie $\nabla_{R_\x}\hat{\cc}(\x)$.  Assume  $\tilde{R}_\x = e^{\delta U}R$, where $\delta U=\left[\begin{array}{ccc} 0 & -\delta\mu_3 & \delta\mu_2\\ \delta\mu_3 & 0 & -\delta\mu_1\\ -\delta\mu_2 & \delta\mu_1 & 0\end{array}\right]$  is a skew-symmetric matrix parameterized by  $\delta\bmu=\left[\begin{array}{ccc} \delta\mu_1 & \delta\mu_2 & \delta\mu_3\end{array}\right]^{\top}$. From this construction, $\delta U$ is the tangent vector at $R_\x$ on the manifold of rotation matrices and $\tilde{R}_\x$ is also a rotation matrix. Based on Taylor expansion, we have 
the first order approximation of $\M(\tilde{R}_\x)\cc^{\atlas}(\x)$ as 
\begin{align*}
\M(\tilde{R}_\x)\cc^{\atlas}(\x) \approx \hat{\cc}(\x) +  \nabla_{R_\x}^\top\hat{\cc}(\x) \delta \bmu \ .
\end{align*}
Hence, we can compute $\nabla_{R_\x}\hat{\cc}(\x)$ as follows.  Assume $\delta U_1, \delta U_2, \delta U_3$ to be skew-symmetric matrices respectively constructed from $[\delta \mu_1, 0, 0]^\top,[0, \delta \mu_2, 0]^\top,[0, 0, \delta \mu_3]^\top$. We have 
\begin{align}
\label{eq:dChat}
\nabla_{R_\x}\hat{\cc}(\x)\approx\left[\begin{array}{c} \left({\bf M}(e^{\delta U_1})\hat{\cc}(\x)-\hat{\cc}(\x)\right)^{\top}/\delta\mu_1 \\ \left({\bf M}(e^{\delta U_2})\hat{\cc}(\x)-\hat{\cc}(\x)\right)^{\top}/\delta\mu_2 \\ \left({\bf M}(e^{\delta U_3})\hat{\cc}(\x)-\hat{\cc}(\x)\right)^{\top}/\delta\mu_3 \end{array}\right].
\end{align} 
It is worth noting that this formulation significantly reduces the computational cost for $\nabla_{R_\x}\hat{\cc}(\x)$. Since $\delta\bmu$ is independent of spatial location $\x$,  ${\bf M}(e^{\delta U_1})$, ${\bf M}(e^{\delta U_2})$, and ${\bf M}(e^{\delta U_3})$ are only calculated once and applied to all $\x$. 

We now compute the variation of $R_\x^\epsilon$ with respect to the small variation of $\phi_1^\epsilon$. This has been referred as exact finite-strain differential that was solved in \cite{Dorst_TPAMI2005} and applied to the DTI tensor-based registration in \cite{Yeo_TMI2009}. Here, we directly adopt the result from \cite{Yeo_TMI2009} and obtain 
\begin{align}
\label{eqn:variationR}
\frac{\partial R_\x^\epsilon}{\partial \epsilon} \Big|_{\epsilon=0} = -F_\x\sum^3_{i=1}\left[\rb_i\times(D_\x h^{\top})_i\right]  \ ,
\end{align}
where $F_\x=-R_\x^{\top}\Big(\tr\left((D_\x\phi_1 D_\x^{\top}\phi_1)^{1/2}\right) \Id-(D_\x\phi_1 D_\x^{\top}\phi_1)^{1/2}\Big)^{-1}R_\x$. $\times$ denotes as the cross product of two vectors.  $(A)_i$ denotes the $i$th column of matrix $A$. $\rb_i=(R_\x^{\top})_i$.  

Given Eq. \eqref{eq:dChat} and \eqref{eqn:variationR}, we thus have 
\begin{align}
\label{eq:termB}
\text{Term (B)} &=-2\int_{\x\in\Omega} \Big\langle\hat{\cc}(\x)\circ \phi_1^{-1}-\cc^{(s)}(\x), \Big( \nabla_{R_\x} \hat{\cc}^\top(\x) F_\x\sum^3_{i=1}\left[\rb_i\times(D_\x h^{\top})_i\right]  \Big)\circ \phi_1^{-1} \Big\rangle d\x \\ \nonumber
&=-2\int_{\x\in\Omega} \ww_\x^{\top}\sum^{3}_{i=1}\left[\rb_i\times(D_\x h^{\top})_i\right] d\x \\ \nonumber
& =-2\int_{\x\in\Omega} \sum^{3}_{i=1}\left\langle \ww_\x \times \rb_i ,\ub_i\right\rangle d\x \ ,
\end{align}
where 
\begin{align}
\label{eq:ww} \ww_\x^{\top}=\bigg(\nabla_{R_\x}\hat{\cc}\Big(\hat{\cc}\big(\x\big)-
\dd\big(\phi_1(\x)\big)\Big)\bigg)^{\top}F_\x\det\big(D_\x\phi_1\big) \ ,
\end{align}
and 
$h=\left[\begin{array}{ccc} h_1 & h_2 & h_3\end{array}\right]^{\top} \ . 
$
$D_\x h$ is approximated as 
$$
D_{\x} h=\left[\begin{array}{c} \nabla_{\x} h_{1}^{\top}\\ \nabla_{\x} h_{2}^{\top}\\ \nabla_{\x} h_{3}^{\top}\end{array}\right]\approx\frac{1}{2\Delta d}\left[\begin{array}{ccc} h_{1,\x^{X+}}-h_{1,\x^{X-}} \;\; & h_{1,\x^{Y+}}-h_{1,\x^{Y-}} \;\;& h_{1,\x^{Z+}}-h_{1,\x^{Z-}}\\ h_{2,\x^{X+}}-h_{1,\x^{X-}} \;\; & h_{2,\x^{Y+}}-h_{2,\x^{Y-}} \;\;& h_{2,\x^{Z+}}-h_{2,\x^{Z-}}\\ h_{3,\x^{X+}}-h_{3,\x^{X-}}\;\; & h_{3,\x^{Y+}}-h_{3,\x^{Y-}} \;\;& h_{3,\x^{Z+}}-h_{3,\x^{Z-}} \end{array}\right],
$$
where $\{\x^{X+},\x^{X-},\x^{Y+},\x^{Y+},\x^{Z+},\x^{Z-}\}$ are the neighbors of $\x$ in $x, y, z$ directions, respectively. $\Delta d$ is the distance of these neighbors to $\x$. Here, term (B) seeks the spatial transformation $\phi_t$ such that the local diffusion profiles of the atlas and subject's HYDIs have to be aligned.

In summary, we have 
\begin{align}
\label{eqn:partialE}
\frac{\partial E}{\partial\epsilon}\Big|_{\epsilon=0}&\approx  -2\int_{\x\in\Omega} \Big\langle \big(D_\x\phi_1\big)^{-\top}\nabla_{\x} \hat{\cc}(\x)\Big(\hat{\cc}(\x)-\cc^{(s)}\big(\phi_1(\x)\big)\Big) \det{\big(D_\x\phi_1\big)}, h \Big\rangle d\x \\ \nonumber
& -\frac{1}{\Delta d} \int_{\x\in\Omega} \sum_{k=1}^{3} \left\{ \left\langle  \ww_{\x} \times \rb_k , \left[\begin{array}{c} h_{k,\x^{X+}}\\ h_{k,\x^{Y+}}\\ h_{k,\x^{Z+}}\end{array}\right] \right\rangle -  \left\langle  \ww_{\x} \times \rb_k , \left[\begin{array}{c}  h_{k,\x^{X-}}\\  h_{k,\x^{Y-}}\\  h_{k,\x^{Z-}}\end{array}\right] \right\rangle \right\} d\x.
\end{align} 
Therefore,  $\nabla_{\phi_1} E$ can be obtained from Eq. \eqref{eqn:partialE}. 

\subsection{Numerical Implementation}
We so far derive $J$ and its gradient $\nabla J(m_t)$ in the continuous setting. In this section, we elaborate the numerical implementation of our algorithm under the discrete setting. Since  HYDI DW signals were represented using the orthonormal BFOR signal bases, both the computation of $J$ in Eq. \eqref{eqn:metric} and the gradient computation in Eq. \eqref{eqn:partialE} do not explicitly involve the calculation $\bPsi(\s)$. Hence, we do not need to discretize the $q$-space. In the discretization of the image domain, we first represent the ambient space, $\Omega$, using a finite number of points on the image grid, $\Omega \cong \{(\x_i)_{i=1}^N\}$. In this setting, we can assume $m_t$ to be the sum of Dirac measures, where $\alpha_i(t)$ is the momentum vector at $\x_i$ and time $t$.
We use a conjugate gradient routine to perform the minimization of $J$ with respect to $\alpha_i(t)$. We summarize steps required in each iteration during the minimization process below: 
\begin{enumerate}
\item Use the forward Euler method to compute the trajectory based on the flow equation:
\begin{eqnarray}
\label{eqn:flowEquation}
\frac{d\phi_t(\x_i)}{dt} &=& \sum_{j=1}^N k_V(\phi_t(\x_i),\phi_t(\x_j)) \alpha_j(t) \ .
\end{eqnarray}
\item Compute $\nabla_{\phi_1(\x_i)} E$ based on Eq. \eqref{eqn:partialE}.
\item Solve $\eta_t =[\eta_i(t)]_{i=1}^N$ in Eq. \eqref{eqn:eta} using the backward Euler integration, where $i$ indices $\x_i$, with the initial condition $\eta_i(1)=\nabla_{\phi_1(\x_i)} E$.
\item Compute the gradient $\nabla J(\alpha_i(t)) = 2 \alpha_i(t) + \eta_i(t)$.
\item Evaluate $J$ when $\alpha_i(t)= \alpha^{\text{old}}_i(t)
- \epsilon \nabla J(\alpha_i(t)) $, where $\epsilon$ is the adaptive step size determined by a golden section search. 
\end{enumerate}

\section{Bayesian HYDI Atlas Estimation}
\label{sec:atlas}

The above registration problem requires an atlas. In this section, we introduce a general framework for Bayesian HYDI atlas estimation. Given $n$ observed HYDI  datasets $\DSI^{(i)}$ for $i=1,\dots, n$, we assume that each of them can be estimated through an unknown atlas $\DSI^{\atlas}$ and a diffeomorphic transformation $\phi^{(i)}$ such that 
\begin{align}
\DSI^{(i)} \approx {\hat{\DSI}}^{(i)}= \phi^{(i)} \cdot \DSI^{\atlas}.
\end{align}
The total variation of $\DSI^{(i)}$ relative to ${\hat{\DSI}}^{(i)}$ is then denoted by $\sigma^2$. The goal here is to estimate the unknown atlas $\DSI^{\atlas}$ and the variation $\sigma^2$. To solve for the unknown atlas $\DSI^{\atlas}$, we first introduce an ancillary ``hyperatlas'' $\DSI_0$, and assume that our atlas is generated from it via a diffeomorphic transformation of $\phi$ such that $\DSI^{\atlas}=\phi \cdot \DSI_0$. We use the Bayesian strategy to estimate $\phi$ and $\sigma^2$ from the set of observations $\DSI^{(i)}, i=1,\dots, n$ by computing the maximum a posteriori (MAP) of $f_{\sigma}(\phi |\DSI^{(1)}, \DSI^{(2)},\dots, \DSI^{(n)}, \DSI_0)$. This can be achieved using the Expectation-Maximization algorithm by first computing the log-likelihood of the complete data ($\phi, \phi^{(i)}, \DSI^{(i)},i=1,2,\dots,n$) when $\phi^{(1)}, \cdots, \phi^{(n)}$ are introduced as hidden variables. We denote this likelihood as $f_{\sigma}( \phi, \phi^{(1)}, \dots, \phi^{(n)}, \DSI^{(1)},\dots \DSI^{(n)}| \DSI_0)$. We consider that the paired information of individual observations, $(\DSI^{(i)}, \phi^{(i)})$ for  $i=1,\dots, n$, as independent and identically distributed. As a result, this log-likelihood can be written as 
\begin{align}
\label{eqn:loglikelihood}
&  \log f_{\sigma}( \phi, \phi^{(1)}, \dots, \phi^{(n)}, \DSI^{(1)},\dots \DSI^{(n)} | \DSI_0) \\ \nonumber
=\; & \log f(\phi | \DSI_0) + \sum_{i=1}^n \Big\{ \log f(\phi^{(i)} | \phi, \DSI_0) + \log f_{\sigma}(\DSI^{(i)}|\phi^{(i)}, \phi, \DSI_0) 
\Big\}  \ ,
\end{align}
where $f(\phi | \DSI_0)$ is the shape prior (probability distribution) of the atlas given the hyperatlas, $\DSI_0$. $f(\phi^{(i)} | \phi, \DSI_0)$ is the distribution of random diffeomorphisms given the estimated atlas ($\phi \cdot \DSI_0$). $f_{\sigma}(\DSI^{(i)}|\phi^{(i)}, \phi, \DSI_0)$ is the conditional likelihood of the HYDI data given its corresponding hidden variable $\phi^{(i)}$ and the estimated atlas ($\phi \cdot \DSI_0$). In the remainder of this section, we first adopt $f(\phi | \DSI_0)$ and $f(\phi^{(i)} | \phi, \DSI_0)$ introduced in \citep{jun_nimg_2008,Qiu:TIP10} under the framework of LDDMM and then describe how to calculate $f_{\sigma}(\DSI^{(i)}|\phi^{(i)}, \phi, \DSI_0)$ in \S \ref{subsec:likelihood} based on the BFOR representation of HYDI.

\subsection{The Shape Prior of the Atlas and the Distribution of Random Diffeomorphisms}

In the framework of LDDMM-HYDI,  we adopt the previous work in \cite{jun_nimg_2008,Qiu:TIP10} and briefly describe the construction of the shape prior (probability distribution) of the atlas, $f(\phi | \DSI_0)$.  Under the LDDMM framework, we can compute the prior $f(\phi |\DSI_0)$ via $m_0$, \ie
\begin{align}
f(\phi|\DSI_0) = f(m_0| \DSI_0) \ ,
\end{align}
where $m_0$ is initial momentum defined in the coordinates of $\DSI_0$ such that it uniquely determines diffeomorphic geodesic flows from $\DSI_0$ to the estimated atlas. When $\DSI_0$ remains fixed, the space of the initial momentum $m_0$ provides a linear representation of the nonlinear diffeomorphic shape space, $\DSI^{\atlas}$, in which linear statistical analysis can be applied. Hence, assuming $m_0$ is random, we immediately obtain a stochastic model for {\it diffeomorphic transformations} of $\DSI_0$. More precisely, we follow the work in \citep{jun_nimg_2008,Qiu:TIP10} and assume $m_0$ to be a centered Gaussian random field (GRF) model. The distribution of $m_0$ is characterized by its covariance bilinear form, defined by
\begin{align*}
\Gamma_{m_0}(v, w) = E\bigl[ m_0(v) m_0(w) \bigr] \ ,
\end{align*}
where $v,w$ are vector fields in the Hilbert space of $V$ with reproducing kernel $k_V$. 

\noindent We associate $\Gamma_{m_0}$ with $k_V^{-1}$. The ``prior'' of $m_0$ in this case is then 
$$\frac{1}{\mathcal{Z}}\exp{\left(-\frac{1}{2}\langle m_0, k_V m_0 \rangle_2\right)},$$
where $\mathcal{Z}$ is the normalizing Gaussian constant. This leads to formally define the ``log-prior'' of $m_0$ to be
\begin{align}
\label{eqn:log-prior}
\log f(m_0 | \DSI_0) \approx -\frac{1}{2} \langle m_0, k_V m_0 \rangle_2  \ ,
\end{align}
where we ignore the normalizing constant term $\log{\mathcal{Z}}$.

We can construct the distribution of random diffeomorphisms, $f(\phi^{(i)} | \phi, \DSI_0)$, in the similar manner. We define $f(\phi^{(i)} | \phi, \DSI_0)$ via the corresponding initial momentum $m_0^{(i)}$ defined in the coordinates of $\phi \cdot \DSI_0$. We also assume that $m_0^{(i)}$ is random, and therefore, we again obtain a stochastic model for {\it diffeomorphic transformations} of $\DSI^{\atlas} \cong \phi \cdot \DSI_0$. $m_0^{(i)}$ is assumed to be a centered GRF model with its covariance as $k_V^\pi$, where $k_V^{\pi}$ is the reproducing kernel of the smooth vector field in a Hilbert space $V$. Hence, we can define the log distribution of random diffeomorphisms as 
\begin{align}
\label{eqn:loglikelihoodrandiff}
 \log f(\phi^{(i)} | \phi, \DSI_0) \approx -\frac{1}{2} \langle m_0^{(i)}, k_V^{\pi} m_0^{(i)} \rangle_2  \ .
\end{align} 
where as before, we ignore the normalizing constant term $\log{\mathcal{Z}}$.

\subsection{The Conditional Likelihood of the HYDI Data} 
\label{subsec:likelihood}

Given the representation of the diffusion signals in the $q$-space using BFOR bases, we construct the conditional likelihood of the HYDI data $f_{\sigma}(\DSI^{(i)}|\phi^{(i)}, \phi, \DSI_0)$ via the BFOR coeffcients. We assume that $\cc_{nj}(\x)$ has a multivariate Gaussian distribution with mean of $\cc^{\atlas}(\x)$ and covariance  $\sigma^2 \Id$, where $\cc^{\atlas}(\x)$ are the BFOR coefficients associated with $\DSI_{\atlas}$. 

From the Gaussian assumption, we can thus write the conditional ``log-likelihood'' of $\DSI^{(i)}$ given $\DSI^{\atlas}$ and $\phi_1^{(i)}$ as 
\begin{align}
\label{eqn:conditionalloglikelihood}
&\log f_{\sigma}(\DSI^{(i)}|\phi_1^{(i)}, \phi_1, \DSI_0)  \\ \nonumber
\approx & \int_{\x \in \bOmega} \Big\{ -\frac{1}{2\sigma^2} \bigg\| \big(\M(R_{\x})\;\cc^{\atlas}(\x)\big)\circ(\phi_1^{(i)})^{-1}(\x)  - \cc^{(i)}(\x)  \bigg\|^2 -\frac{\log \sigma^2}{2} \Big\} d\x  \ ,
\end{align}
where as before, we ignore the normalizing Gaussian term.

\subsection{Expectation-Maximization Algorithm}
\label{subsec:em}

We have shown how to compute the log-likelihood shown in Eq. \eqref{eqn:loglikelihood}. In this section, we will show how we employ the Expectation-Maximization algorithm to estimate the atlas, $\DSI^{\atlas}(\x,\s)$, for $\s \in \Do^2, \x\in \bOmega$, and $\sigma^2$. From the above discussion, we first rewrite the log-likelihood function of the complete data in Eq. \eqref{eqn:loglikelihood} as 
\begin{align}
&   \log f_{\sigma}( m_0, m_0^{(1)}, \dots, m_0^{(n)}, \DSI^{(1)},\dots \DSI^{(n)} | \DSI_0)  \\ \nonumber
\approx& -\frac{1}{2} \langle m_0, k_V m_0 \rangle_2  -\sum_{i=1}^n \Bigg\{ \frac{1}{2} \langle m_0^{(i)}, k_V^{\pi} m_0^{(i)} \rangle_2 \\ \nonumber 
&+   
   \int_{\x \in \bOmega} \Big\{ \frac{1}{2\sigma^2} \bigg\| \big(\M(R_{\x})\;\cc^{\atlas}(\x)\big)\circ(\phi_1^{(i)})^{-1}(\x)  - \cc^{(i)}(\x)  \bigg\|^2  +\frac{\log \sigma^2}{2} \Big\} d\x
\Bigg\} \ ,
\end{align}
where $\DSI^{\atlas}(\x,\s) = \phi_1 \cdot \DSI_0(\x,\s)$ and can be computed based on Eq. \eqref{eqn:diffeoaction}.

\bigskip \noindent\textbf{The E-Step.} The E-step computes the expectation of the complete data log-likelihood given the previous atlas $m_0^{\old}$ and variance ${\sigma^{2}}^{\old}$. We denote this expectation as 
$Q(m_0, \sigma^2|m_0^{\old},{\sigma^{2}}^{\old})$ given in the equation below,
\begin{align}
\label{eqn:qfun}
& Q\left(m_0, \sigma^2|m_0^{\old},{\sigma^{2}}^{\old}\right) \\ \nonumber
= &E\Bigg\{ \log f_{\sigma}( m_0, m_0^{(1)}, \dots, m_0^{(n)}, \DSI^{(1)},\dots \DSI^{(n)} | \DSI_0) \Big| m_0^{\old},{\sigma^{2}}^{\old}, \DSI^{(1)}, \cdots,  \DSI^{(n)}, \DSI_0 \Bigg\} \\ \nonumber
 \approx &-\frac{1}{2} \langle m_0, k_V m_0 \rangle_2 - \sum_{i=1}^n E\Bigg[ \frac{1}{2} \langle m_0^{(i)}, k_V^{\pi} m_0^{(i)} \rangle_2  \\ \nonumber
+ &   \int_{\x \in \bOmega} \Big\{ \frac{1}{2\sigma^2} \bigg\| \big(\M(R_{\x})\;\cc^{\atlas}(\x)\big)\circ(\phi_1^{(i)})^{-1}(\x)  - \cc^{(i)}(\x)  \bigg\|^2 +\frac{\log \sigma^2}{2} \Big\} d\x
\Bigg] \ .
\end{align}

\bigskip \noindent\textbf{The M-Step.} The M-step generates the new atlas by maximizing the $Q$-function with respcet to $m_0$ and $\sigma^2$.   The update equation is given as  
\begin{align}
\label{eqn:qopt}
 & m_0^{\new}, {\sigma^2}^{\new} \\ \nonumber
=  &\argmax_{m_0,\sigma^2} Q\left(m_0, \sigma^2 |m_0^{\old}, {\sigma^2}^{\old} \right) \\ \nonumber
= &\argmin_{m_0,\sigma^2} \Bigg\{\langle m_0, k_V m_0\rangle_2 \\ \nonumber
& + \sum_{i=1}^n  E\Bigg[ \int_{\x \in \bOmega} \Big\{ \frac{1}{\sigma^2} \bigg\| \big(\M(R_{\x})\;\cc^{\atlas}(\x)\big)\circ(\phi_1^{(i)})^{-1}(\x)  - \cc^{(i)}(\x)  \bigg\|^2+\log \sigma^2 \Big\} d\x \Bigg] \Bigg\}   \ ,
\end{align}
where we use the fact that the conditional expectation of $\langle m_0^{(i)}, k_V^\pi m_0^{(i)}\rangle_2 $ is constant. We solve $\sigma^2$ and $m_0$ by separating the procedure for updating $\sigma^2$ using the current value of $m_0$, and then optimizing $m_0$ using the updated value of $\sigma^2$. 

We now derive how to update values of $\sigma^2$ and $m_0$ from ${\it {Q}}$-function in Eq. \eqref{eqn:qopt}. It is straightforward to obtain $\sigma^2$ by taking the derivative of $Q\left(m_0, \sigma^2 | m_0^{\old}, {\sigma^2}^{\old}\right)$ with respect to $\sigma^2$ and setting it to zero. Hence, we have 
\begin{align}
\label{eqn:updatesigma}
&{\sigma^2}^{\new} = \frac{1}{n} \sum_{i=1}^{n}   \int_{x \in \bOmega} \bigg\| \big(\M(R_{\x})\;\cc^{\atlas}(\x)\big)\circ(\phi_1^{(i)})^{-1}(\x)  - \cc^{(i)}(\x)  \bigg\|^2 d\x      \ . 
\end{align}

For updating $m_0$, let $y=\left(\phi_1^{(i)}\right)^{-1}(\x)$. Using the change of variables strategy, we have 
{\allowdisplaybreaks
\begin{align*}
&\sum_{i=1}^n E\Bigg[ \int_{x \in \bOmega} \frac{1}{2\sigma^2} \bigg\| \big(\M(R_{\x})\;\cc^{\atlas}(\x)\big)\circ(\phi_1^{(i)})^{-1}(\x)  - \cc^{(i)}(\x)  \bigg\|^2  d\x\Bigg]\\
=&\sum_{i=1}^n E\Bigg[ \int_{\y \in \bOmega} 
\frac{1}{2\sigma^2}  \bigg\| \M(R_{\y})\;\cc^{\atlas}(\y)  - \cc^{(i)}(\y)\circ \phi_1(\y)  \bigg\|^2  |D\phi_1^{(i)}(\y)|d\y
   \Bigg]\\
=&\int_{y \in \bOmega} 
\frac{1}{2\sigma^2}
 \sum_{i=1}^n E\Bigg[ \bigg\|\M(R_{\y})\;\cc^{\atlas}(\y)  - \overline{\cc}_0(\y) + \overline{\cc}_0(\y) -  \cc^{(i)}(\y)\circ \phi_1(\y) \bigg\|^2   \Bigg]   d\y\\
=&\int_{y \in \bOmega}  
\frac{1}{2\sigma^2}
\sum_{i=1}^n E\Bigg[  \Big\{ \Big\|\M(R_{\y})\;\cc^{\atlas}(\y)  - \overline{\cc}_0(\y) \Big\|^2 + \Big\|\overline{\cc}_0(\y) -  \cc^{(i)}(\y)\circ \phi_1(\y) 
\Big\|^2  \\ \nonumber 
& +2 \Big\langle \M(R_{\y})\;\cc^{\atlas}(\y)  - \overline{\cc}_0(\y) , \overline{\cc}_0(\y) -  \cc^{(i)}(\y)\circ \phi_1(\y)  \Big\rangle \Big\} |D\phi_1^{(i)}(\y)| \Bigg] d\y \ .
\end{align*}}

Since the second item in the above equation is independent of $m_0$ and $\sum_{i=1}^n |D\phi_1^{(i)}(\y)|\big( \overline{\cc}_0(\y) -  \cc^{(i)}(\y)\circ \phi_1(\y) \big) =0 $ ,we have
\begin{align}
\label{eqn:updatemomenta}
m_0^{\new}  =  \argmin_{m_0} \left\{ \langle m_0, k_V m_0\rangle_2 + \frac{1}{\sigma^{2new}} \int_{\y \in \bOmega} \alpha(\y) \bigg\| \big(\M(R_{\y})\;\cc_0(\y)\big)\circ \phi_1^{-1}(\y)  - \overline{\cc}_0(\y)  \bigg\|^2 d\y   \right\}\ ,
\end{align}
where $\alpha(\y) = \sum_{i=1}^n |D\phi_1^{(i)}(\y)| $ and $\displaystyle|D\phi_1^{(i)}|$ is the Jacobian determinant of $\phi_1^{(i)}$. $\overline{\cc}_0(\x)$ is given as 
 \begin{align}
\label{eqn:I01}
\overline{\cc}_0(\x)=\frac{1}{\sum_{j=1}^n |D\phi_1^{(j)}(\x)|}
\sum_{i=1}^n |D\phi_1^{(i)}(x)| \big(\M(R_{\x})\;\cc^{(i)}(\x)\big)\circ(\phi_1^{(i)})(\x) \ .
 \end{align}
 The variational problem listed in Eq. \eqref{eqn:updatemomenta} is referred as ``modified LDDMM-HYDI mapping", where weight $\alpha$ is introduced. It can be easily solved by adopting the LDDMM-HYDI algorithm in Section \ref{sec:LDDMM-HYDI}. We present the steps involved in the EM optimization in Algorithm \ref{alg:em}. 

\begin{algorithm}
\caption{\label{alg:em}\bf (The EM Algorithm for the HYDI Atlas Generation)}
We initialize $m_0=0$. Thus, the hyperatlas $\DSI_0$ is considered as the initial atlas. 
\begin{enumerate}
\item Apply the LDDMM-HYDI mapping algorithm in Section \ref{sec:LDDMM-HYDI} to register the current atlas to each individual HARDI dataset, which yields $m_0^{(i)}$ and $\phi_t^{(i)}$.
\item Compute $\overline{\cc}_0 $ according to Eq. \eqref{eqn:I01}.
\item Update $\sigma^2$ according to Eq. \eqref{eqn:updatesigma}.
\item Estimate $\DSI^{\atlas} = \phi_1 \cdot \DSI_0$, where $\phi_t$ is found by applying 
the modified LDDMM-HYDI mapping algorithm as given in Eq. \eqref{eqn:updatemomenta}.
\end{enumerate}
The above computation is repeated until the atlas converges.
\end{algorithm}

\section{Experiments}
\label{sec:expts}
In this section, we show the atlas generated using Algorithm \ref{alg:em}, evaluate the LDDMM-HYDI mapping accuracy, and compare it with that of an existing method, Advanced Normalization Tools (ANTs) \cite{Ants_Avants2008}. 
The experiments were performed on 36 healthy, adult human brain datasets (age: $61.8 \pm 6.47$ years) acquired using a 3.0T GE-SIGNA scanner with an 8-channel head coil and ASSET parallel imaging. The $in$ $vivo$ datasets were acquired using a hybrid, non-Cartesian sampling scheme \cite{Wu_Alexander_HYDI}, and the HYDI encoding scheme is described in Table \ref{table:Encoding}. Since EAP reconstruction is sensitive to angular resolution, the number of encoding directions is increased with each shell to increase the angular resolution with the level of diffusion weighting. The number of directions in the outer shells were increased to better characterize complex tissue organization. In the experiments of the atlas generation and LDDMM-HYDI evaluation, we represented HYDI DW signals using the BFOR signal basis with upto the fourth order modified SH bases and upto the sixth order spherical Bessel functions. The corresponding BFOR expansion coefficients were used in the atlas generation and LDDMM-HYDI optimization.

\begin{table}
\centering
\caption{HYDI encoding scheme for human datasets.}
\begin{tabular}{ l   l   l   l   l }
\hline
Shell & Ne & $q$ (mm$^{-1}$) & $\Delta q$ (mm$^{-1}$) & $b$ (s/mm$^{2}$) \\ \hline
& 7 & 0 & & 0 \\ 
1st & 6 & 15.79 & 15.79 & 300 \\ 
2nd & 21 & 31.58 & 15.79 & 1200 \\ 
3rd & 24 & 47.37 & 15.79 & 2700 \\ 
4th & 24 & 63.16 & 15.79 & 4800 \\ 
5th & 50 & 78.95 & 15.79 & 7500 \\ 
    & Total=132 & $q_{max}$=78.95 & Mean=15.79 & $b_{max}$=7500 \\ \hline
\end{tabular}
\label{table:Encoding}
\end{table}

\subsection{HYDI Atlas}
\label{subsec:HYDI_Atlas}
In the experiment of the atlas generation, we first chose one HYDI dataset as hyperatlas and assumed $m_0=0$ such that the hyperatlas was used as the initial atlas. We then followed Algorithm \ref{alg:em} and repeated it for ten iterations. Notice that $k_V$ associated with the covariance of $m_0$ and $k_V^\pi$ associated with the covariance of $m_0^{(i)}$ were known and predetermined. Since we were dealing with vector fields in $\Re^3$, the kernel of $V$ is a matrix kernel operator in order to get a proper definition. Making an abuse of notation, we defined $k_V$ and $k_V^\pi$ respectively as $k_V \Id_{3 \times 3}$ and $k_V^\pi \Id_{3 \times 3}$,  where $\Id_{3 \times 3}$ is a $3 \times 3$ identity matrix and $k_V$ and $k_V^\pi$ are scalars.  In particular, we assumed that $k_V$ and $k_V^\pi$ are Gaussian with kernel sizes of $\sigma_V$ and $\sigma_{V^\pi}$.
Since $\sigma_V$ determines the smoothness level of the mapping from the hyperatlas to the blur averaged atlas whereas $\sigma_{V^\pi}$ determines that from the sharp atlas to individual HYDI datasets, $\sigma_V$ should be greater than $\sigma_{V^\pi}$. We experimentally determined $\sigma_{V^\pi}=10$ and $\sigma_V=12$. 

\begin{figure}
\centering
\includegraphics[width=0.75\linewidth]{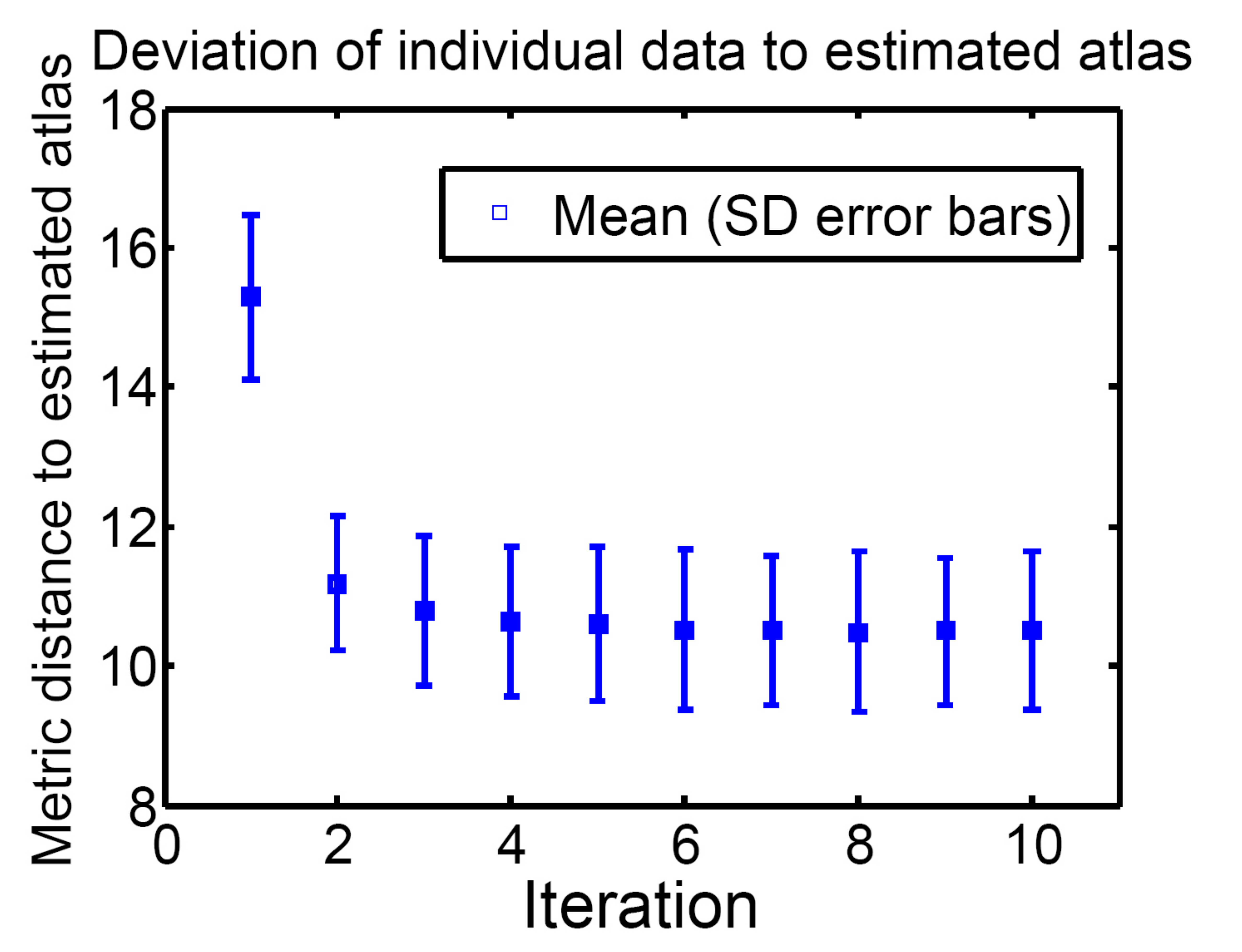}
\caption{The mean and standard deviation of the diffeomorphic metric between individual subjects and the estimated atlas in each iteration.}
\label{fig:optdiffeometric}
\end{figure}

First, we empirically demonstrate the convergence of the diffeomorphic metric between individual subjects and the estimated atlas. This is measured using the square root of the inner product of the initial momentum. Figure \ref{fig:optdiffeometric} shows the mean diffeomorphic metric of individual subjects referenced to the estimated atlas as well as its standard deviation across the subjects. From Figure \ref{fig:optdiffeometric}, we see that the average diffeomorphic metric changed less than $5\%$ after two iterations. 

\begin{figure}
\centering
\includegraphics[width=1\linewidth]{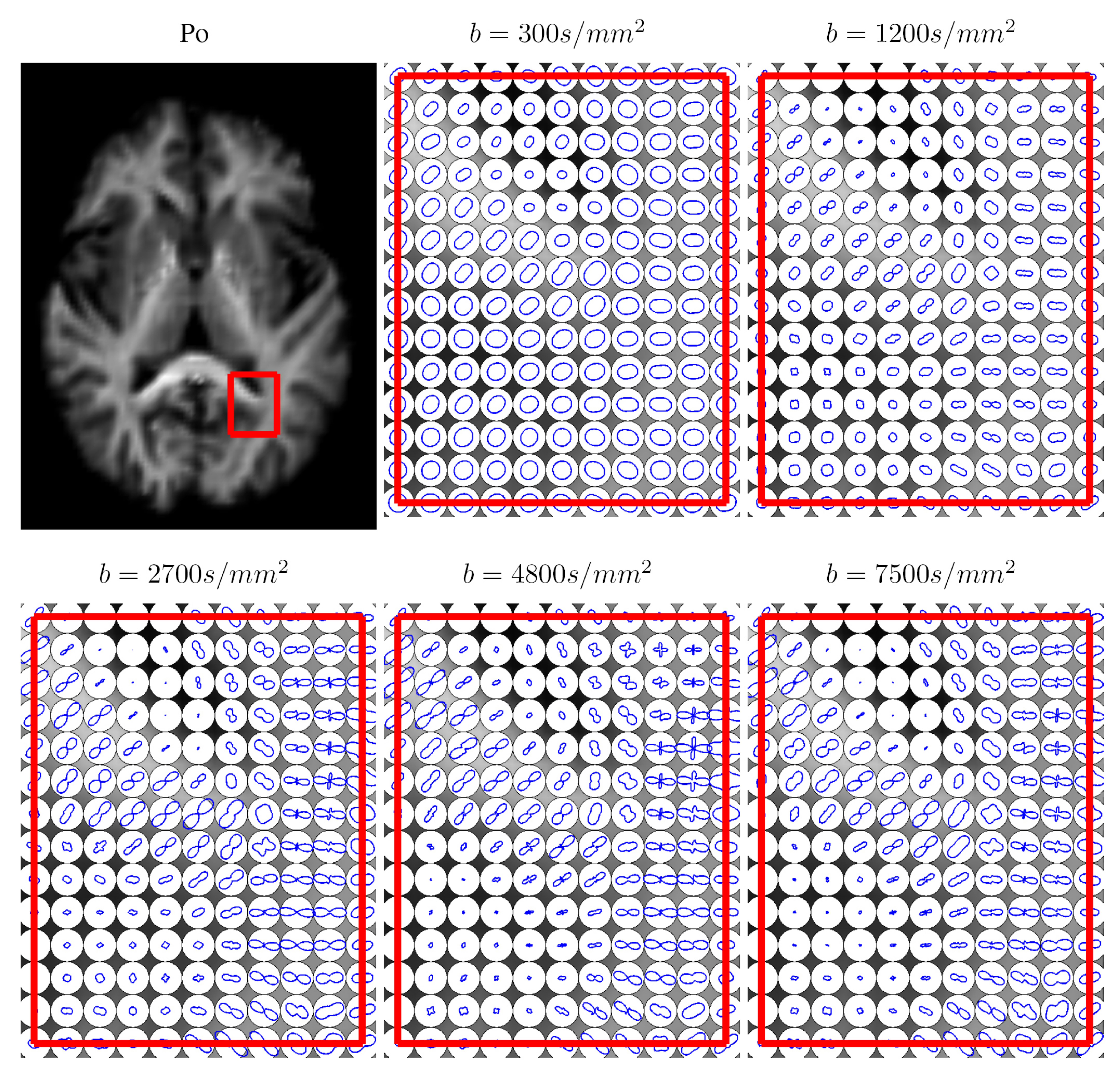}
\caption{Illustration of the estimated HYDI atlas. The first panel shows the atlas of the zero displacement probability (Po) image in the axial view. The rest panels show the diffusion profiles in the region of interest (red frame) at individual shells with $b$=300, 1200, 2700, 4800, and 7500 $s/mm^2$ in the $q$-space. Note that the profile of the diffusion weighted signals is orthogonal to the fiber orientation.}
\label{fig:qshell}
\end{figure}

\begin{figure}
\centering
\includegraphics[width=1\linewidth]{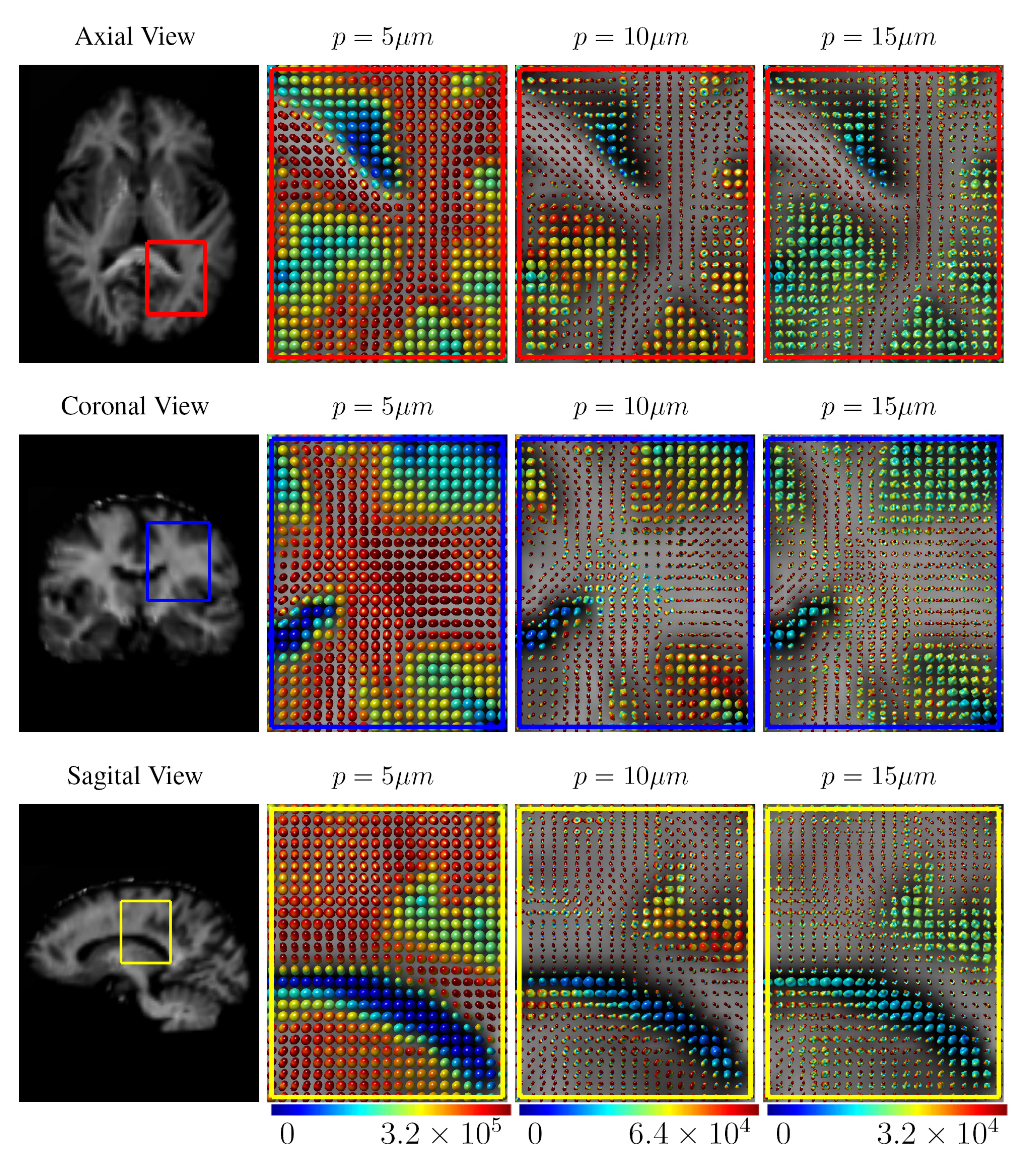}
\caption{The estimated HYDI atlas in the ensemble average propagator (EAP) space. The first column shows the atlas in terms of the zero displacement probability (Po). The second to forth columns respectively illustrate the diffusion profiles of the atlas at three given radii ($p=5,10,15\mu m$) in the EAP space. The color indicates the values of EAP.}
\label{fig:EAPatlas}
\end{figure}

\begin{figure}
\centering
\includegraphics[width=1\linewidth]{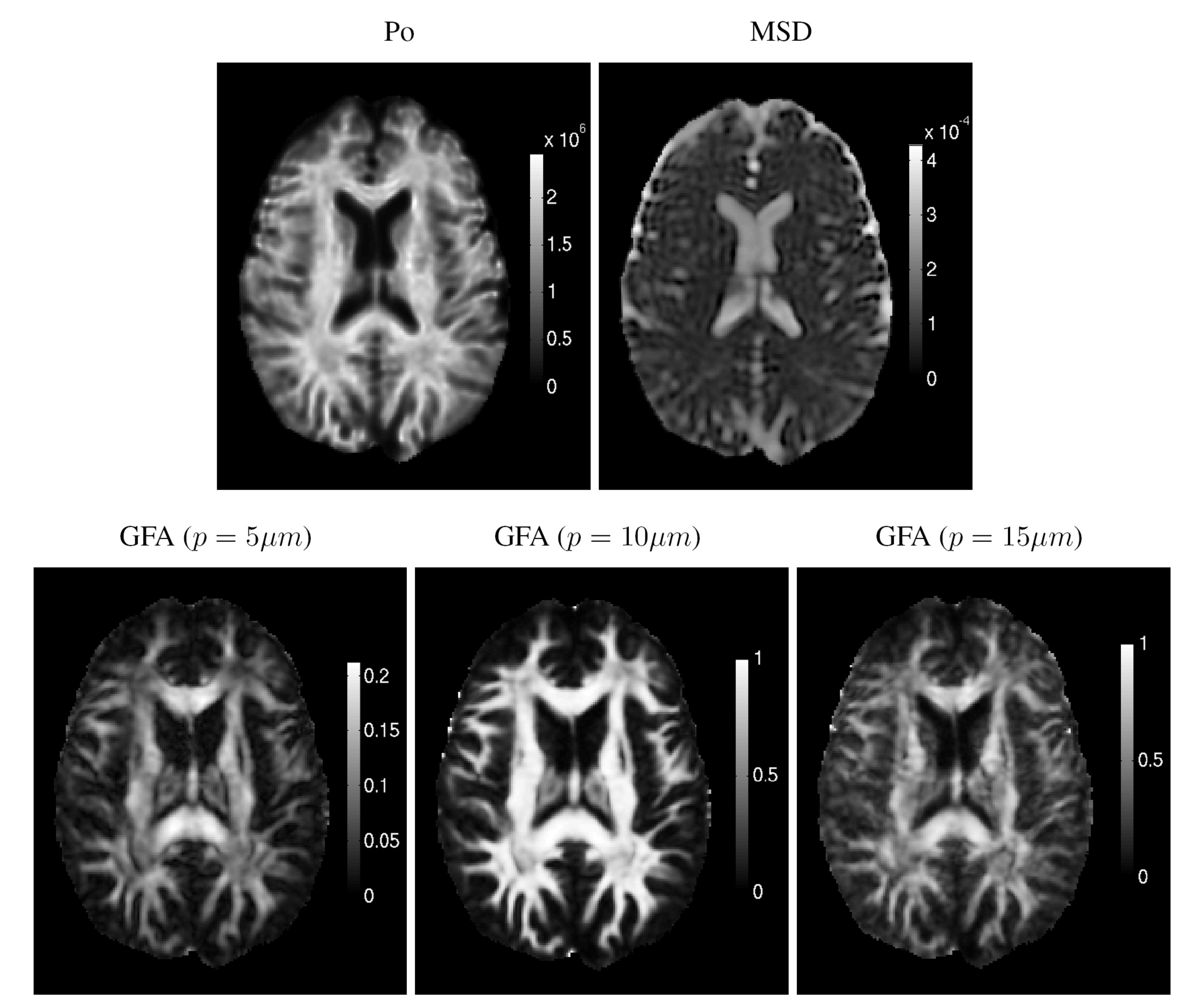}
\caption{Illustration of the estimated altas on the zero-displacement probability (Po), mean squared displacement (MSD), and generalized fractional anisotropy (GFA).}
\label{fig:EAPfeature}
\end{figure}

Next, we illustrate the atlas estimated from the $36$ adults' HYDI datasets after ten iterations. Figure \ref{fig:qshell} illustrates the estimated atlas of the diffusion signals at shells of $b=300, 1200, 2700, 4800, 7500 s/mm^2$. Figure \ref{fig:EAPatlas} shows the reconstructed EAP based on the coefficients of HYDI. Each row respectively shows the estimated atlas in the axial, coronal, and sagital views, while each column shows the zero displacement probability (Po) image derived from EAP and the diffusion profiles of this atlas at three layers of the EAP space. Figure \ref{fig:EAPfeature} demostrates the estimated altas on the zero-displacement probability (Po), mean squared displacement (MSD), and generalized fractional anisotropy (GFA) under three different radii, as introduced in \cite{Pasha_NI2012}. Visually, these figures show that the estimated atlas has the anatomical details of the brain white matter.

\newpage
\subsection{HYDI mapping}
\label{sec:HYDI_mapping}
Given the altas generated in \S \ref{subsec:HYDI_Atlas}, we mapped the 36 subjects into the atlas space using LDDMM-HYDI with $\sigma_V=5$. To evaluate the mapping results, we first illustrate the mapping results of HYDI datasets using LDDMM-HYDI and then evaluate the influence of the reorientation on the optimization of the diffeomorphic transformation, which is often neglected in existing DWI-based registration algorithms (e.g., \cite{Dhollander_MICCAI2010,Dhollander_MICCAI2011}). Finally, we compare the mapping accuracy of LDDMM-HYDI with that of an existing registration method, Advanced Normalization Tools (ANTs) \cite{Ants_Avants2008}. 

\begin{figure}[htb]
\centering
\includegraphics[width=1\linewidth]{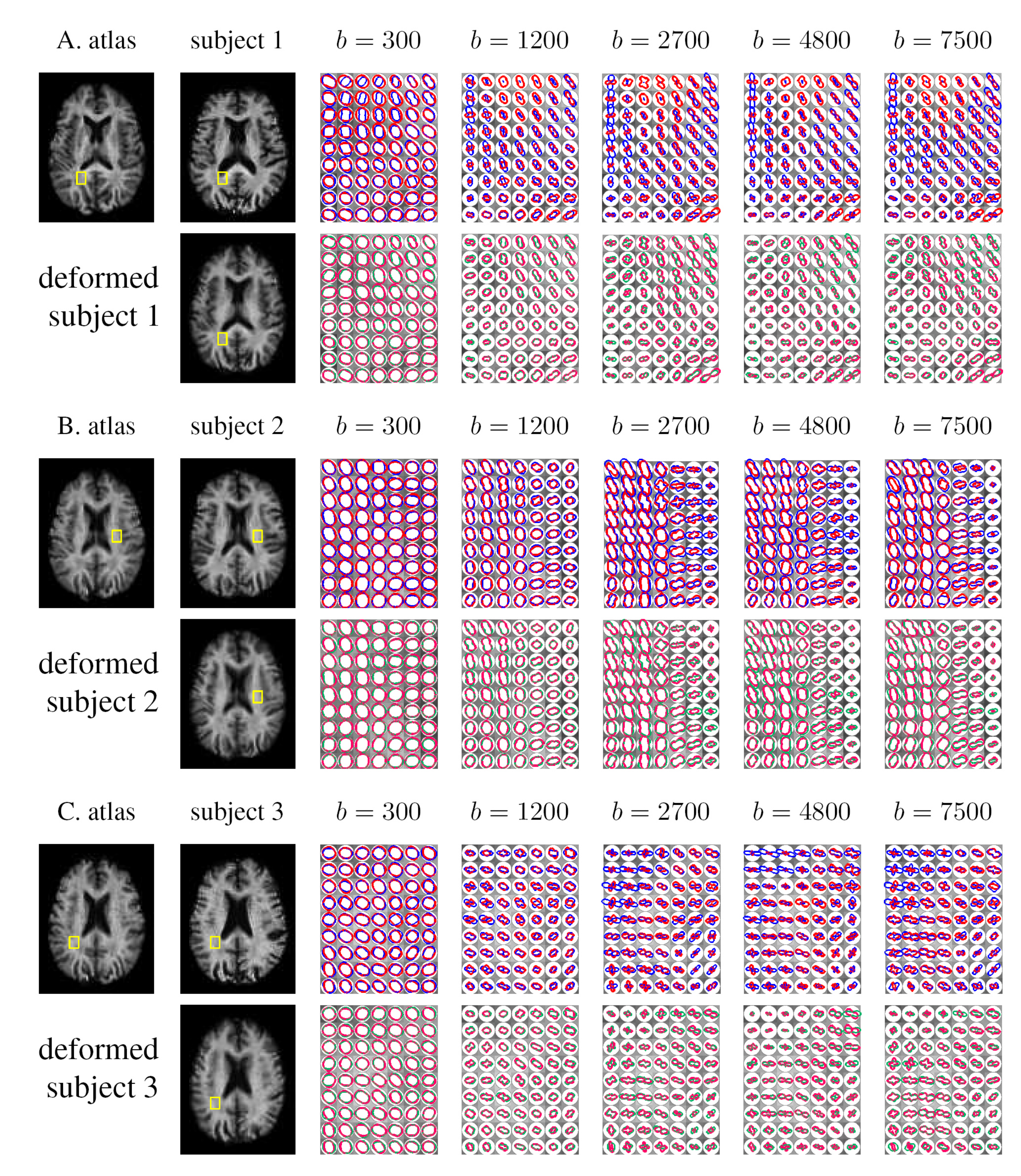}
\caption{Illustration of the LDDMM-HYDI mapping results. The first row of panels (A-C) illustrates the atlas image, subject image, the diffusion profiles at individual shells with $b$=300, 1200, 2700, 4800, and 7500 $s/mm^2$ in $q$-space, respectively. The second row of panels (A-C) illustrates the deformed subject image in atlas space after the LDDMM-HYDI mapping, the diffusion profiles at individual shells with $b$=300, 1200, 2700, 4800, and 7500 $s/mm^2$ in $q$-space, respectively. Red, blue, and green contours in the last five columns respectively illustrate the diffusion profiles of the atlas, subject, and deformed subject in atlas space. The closer the green contour to the red contour, the better the alignment. Note that the profile of diffusion weighted signals is shown in this figure. It is orthogonal to the fiber orientation.}
\label{fig:realresults}
\end{figure}

Figure \ref{fig:realresults} shows the LDDMM-HYDI mapping results of three subjects. The last five columns respectively illustrate the geometric shapes of the diffusion signals at five shells of the $q$-space in the brain regions with crossing fibers. Red, blue, and green contours respectively represent the shape of the diffusion signals from the atlas, subject, and deformed subjects. Visually, the diffusion profiles at each shell can be matched well after the mapping. 

We next evaluated the mapping accuracy of the LDDMM-HYDI algorithms with and without the computation of Term (B) in Eq. \eqref{eq:Egrad} during the optimization, where Term (B) seeks the diffeomorphic transformation such that the local diffusion profiles of the atlas and subject's HYDIs can be aligned. For this, we first computed the diffusion probability density functions (PDFs) of water molecules, \ie the ensemble average propagator (EAP), using Fourier transform \cite{Pasha_NI2012}. Then, we calculated the symmetrized Kullback-Leibler (sKL) divergence between the deformed subject and atlas PDFs \cite{Chiang:TMI08} in major white matter tracts. The smaller sKL metric indicates the better alignment between the deformed subject and atlas HYDIs. The major white matter tracts evaluated in this study include corpus callosum (CC), corticospinal tract (CST), internal capsule (IC),  corona radiata (CR), external capsule (EC), cingulum (CG), superior longitudinal fasciculus (SLF), and  inferior fronto-occipital fasciculus (IFO). Table \ref{tab:real_result_grad} lists the values of the mean and standard deviation of the sKL metric for each major white matter tract among $36$ subjects when the LDDMM-HYDI algorithms with and without the Term (B) computation were respectively employed. Pairwise Student $t$-tests suggest that the LDDMM-HYDI algorithm with the explicit orientation optimization (Term (B) computation) significantly improves the alignment in the major white matter tracts when compared to that without the explicit orientation optimization ($p<0.05$).

Last, we compared the mapping accuracy of LDDMM-HYDI with that of Advanced Normalization Tools (ANTs) \cite{Ants_Avants2008}. The ANTs transformation was found based on Po images. In the ANTs mapping, cross correlation was used to quantify the similarity between the atlas and subject's images. The Gaussian smoothing kernel was set as  $\sigma=5$ and the symmetric regulation weight was $0.01$. The transformation obtained from ANTs was then applied to DWI signals based on reorientation scheme given in Eq. (\ref{eqn:diffeoaction}). Table \ref{tab:real_result_shells} lists the squared difference in the diffusion signals of the atlas and deformed subjects after ANTs and LDDMM-HYDI mapping at individual shells in the $q$-space. Pairwise Student $t$-tests suggested the significant improvement in the alignment of DWIs using LDDMM-HYDI against ANTs ($p<0.05$) at every shell of the $q$-space.


\begin{table}[h]
\caption{Table lists the mean and standard deviation values of the symmetrized Kullback-Leibler (sKL) divergence of the diffusion probability density functions (PDFs) between the deformed subject and atlas HYDIs in each major white matter tract. The second and third columns show the results obtained from the LDDMM-HYDI with and without the Term (B) computation. * denotes statistical significance, suggesting that the alignment obtained from the LDDMM-HYDI with the Term (B) computation is better than that obtained from the LDDMM-HYDI without the Term (B) computation at a significance level of $0.05$. Abbreviations: CC-corpus callosum; CST-corticospinal tract; IC- internal capsule; CR-corona radiata; EC-external capsule, CG-cingulum, SLF-superior longitudinal fasciculus, and  IFO-inferior fronto-occipital fasciculus.}
\begin{center}
\begin{tabularx}{1\textwidth}{|X|c|c|}
\hline
 & LDDMM-HYDI with Term (B) & LDDMM-HYDI without Term (B)\\
\hline
CST & 0.434(0.066) & 0.477(0.084)* \\
CC & 0.346(0.046) & 0.369(0.046)* \\
IC & 0.377(0.044) & 0.387(0.045)* \\
CR & 0.282(0.038) & 0.289(0.038)* \\
EC & 0.322(0.032) & 0.327(0.032)* \\
CG & 0.406(0.048) & 0.417(0.048)* \\
SLF & 0.327(0.060) & 0.341(0.059)* \\
IFO & 0.366(0.037) & 0.374(0.037)* \\ 
\hline 
\end{tabularx}
\end{center}
\label{tab:real_result_grad}
\end{table}

\begin{table}[h]
\caption{Comparison of the mapping accuracy between LDDMM-HYDI and ANTs. The first column lists the b value ($\text{s/mm}^2$) of each shell. The second and third columns list the squared difference in the diffusion signals of the deformed subjects and the atlas after the ANTs and LDDMM-HYDI mapping, respectively. Mean and standard deviation are given. * denotes statistical significance, suggesting that the alignment obtained from LDDMM-HYDI is better than that obtained from ANTs at a significance level of $0.05$.
}
\vspace{-0.4cm}
\begin{center}
\begin{tabular}{| l | c | c |}
\hline
  & ANTs ($\times 10^4$) & LDDMM-HYDI ($\times 10^4$)\\
\hline
b=300  &  5.765(0.488) & 3.236(0.394)* \\
b=1200 &  2.702(0.200) & 1.928(0.183)* \\
b=2700 &  1.145(0.071) & 0.827(0.061)* \\
b=4800 &  0.637(0.041) & 0.462(0.035)* \\
b=7500 &  0.352(0.030) & 0.248(0.024)* \\ 
\hline 
\end{tabular}
\end{center}
\label{tab:real_result_shells}
\end{table}

\section{Conclusion}
In conclusion, we proposed the LDDMM-HYDI variational problem and the Bayesian atlas estimation model based on the BFOR signal basis representation of DWIs. We derived the gradient of this variational problem with the explicit computation of the mDWI reorientation and provided a numeric algorithm without a need of the discretization in the $q$-space. Additionally, we derived the EM algorithm for the estimation of the atlas in the Bayesian framework. Our results showed that 1) the atlas generated contains anatomical details of the white matter anatomy; 2) the explicit orientation optimization is necessary as it improves the alignment of the diffusion profiles of HYDI datasets; 3) the comparison with ANTs suggests the importance for incorporating the full information of HYDI for the multi-shell DWI registration.

\section*{Acknowledgments}
\addcontentsline{toc}{section}{Acknowledgments}
The work was supported by the Young Investigator Award at the National University of Singapore (NUSYIA FY10 P07), the National University of Singapore MOE AcRF Tier 1, Singapore Ministry of Education Academic Research Fund Tier 2 (MOE2012-T2-2-130), and NIH grants (MH84051, HD003352, AG037639, and AG033514).

\section*{Reference}
\bibliographystyle{elsarticle-harv}
\bibliography{intrinsic,dti,ieeetmi_hardireg,eap}
\label{journalend}

\end{document}